\documentclass{article}

\usepackage{PRIMEarxiv}

\usepackage[utf8]{inputenc} 
\usepackage[T1]{fontenc}    
\usepackage{hyperref}       
\usepackage{url}            
\usepackage{booktabs}       
\usepackage{amsfonts}       
\usepackage{nicefrac}       
\usepackage{microtype}      
\usepackage{lipsum}
\usepackage{fancyhdr}       
\usepackage{graphicx}       
\usepackage{amsmath}
\graphicspath{{media/}}     
\usepackage{subcaption}
\usepackage{placeins} 
\title{Explainable AI through a Democratic Lens: DhondtXAI for Proportional Feature Importance Using the D’Hondt Method
}

\author{
  Türker Berk DÖNMEZ\\
  Biomedical Engineering \\
  Sakarya University of Applied Sciences\\
  Sakarya, 54050, TURKIYE\\
  \texttt{turkerberkdonmez@yahoo.com} \\
}

\begin{document}
\maketitle

\begin{abstract}
In democratic societies, electoral systems play a crucial role in translating public preferences into political representation. Among these, the D’Hondt method is widely used to ensure proportional representation, balancing fair representation with governmental stability. Recently, there has been a growing interest in applying similar principles of proportional representation to enhance interpretability in machine learning, specifically in Explainable AI (XAI). This study investigates the integration of D’Hondt-based voting principles in the DhondtXAI method, which leverages resource allocation concepts to interpret feature importance within AI models. Through a comparison of SHAP (Shapley Additive Explanations) and DhondtXAI, we evaluate their effectiveness in feature attribution within CatBoost and XGBoost models for breast cancer and diabetes prediction, respectively. The DhondtXAI approach allows for alliance formation and thresholding to enhance interpretability, representing feature importance as seats in a parliamentary view. Statistical correlation analyses between SHAP values and DhondtXAI allocations support the consistency of interpretations, demonstrating DhondtXAI’s potential as a complementary tool for understanding feature importance in AI models. The results highlight that integrating electoral principles, such as proportional representation and alliances, into AI explainability can improve user understanding, especially in high-stakes fields like healthcare.
\end{abstract}


\section{Introduction}
In democratic societies, election systems play a vital role in translating public preferences into political representation. These systems range from majority and proportional systems to mixed methods, each with unique mechanisms and impacts on representation \cite{ansolabehere2010measuring}. Majority systems tend to favor larger parties and provide stable governments, whereas proportional systems, such as the D'Hondt method, aim to reflect the diversity of voter support across multiple parties. This proportional representation method, while intended to provide a balanced view of political preferences, can sometimes present complexities in accurately measuring democratic outcomes, as factors like district size and voter distribution affect representation fairness \cite{nahuddin2017pemilihan}. Therefore, understanding election systems is essential for assessing how democracies work and how well they reflect the population's intent.

The D'Hondt method, widely used in proportional representation systems, allocates seats to parties based on the number of votes they receive, adjusted by a divisor sequence to ensure representation proportional to each party's share of votes \cite{guney2018mixed}. Its fairness is often debated, as some argue it slightly favors larger parties due to the nature of seat allocation, which can contribute to governmental stability by reducing the fragmentation of legislative bodies \cite{yilmazevaluation}. On the other hand, studies on auditing, such as Stark et al. (2014)\cite{stark2014verifiable}, indicate that the D'Hondt method offers advantages in transparency and verifiability, especially when risk-limiting audits are applied, ensuring election accuracy and fairness. This balance between stability and fair representation is a core consideration in evaluating the suitability of the D'Hondt method for different political contexts.

Countries across Europe and beyond utilize the D'Hondt method to various extents, including Spain, Portugal, and Poland, where it has been integral in ensuring proportional representation within parliamentary systems \cite{boratyn2019formal}. The method's influence extends to both national and regional elections, as observed in nations like Turkey, which employs the D'Hondt method with a national threshold to balance between fair representation and governmental effectiveness \cite{yilmazevaluation}. This widespread adoption underscores the method's adaptability and its perceived fairness in representing diverse political voices while maintaining governance stability.

Artificial intelligence (AI) has significantly evolved, with neural networks and tree-based models becoming central techniques in various domains. Neural networks, such as multilayer perceptrons (MLPs), excel at handling complex, non-linear relationships by adjusting their layered connections between neurons, a structure that enables robust pattern recognition in tasks like disease prediction and environmental modeling \cite{yasin2023multilayer, barbosa2022individual}. Tree-based models, like decision trees and random forests, offer a simpler yet effective structure by segmenting data into branches based on feature values. While tree-based models provide transparent, interpretable decision paths, neural networks remain powerful for capturing intricate relationships, albeit with a trade-off in interpretability. Studies comparing these approaches, like Sorano et al. \cite{sorano2024evolutionary}, highlight that while neural networks often achieve higher accuracy, tree-based models are preferable in applications where clarity in model decisions is essential, underscoring the practical considerations when choosing between these methods.

Explainable AI (XAI) has emerged to address the interpretability of complex AI models, aiming to make model decisions more understandable for human users. Traditional AI models, especially deep neural networks, operate as “black boxes,” with decision processes often opaque to end users. Techniques like SHAP (Shapley Additive Explanations) and LIME (Local Interpretable Model-agnostic Explanations) offer post-hoc explanations by highlighting the influence of individual features on predictions, making AI more accessible in fields such as healthcare and finance, where accountability is critical \cite{dwivedi2023explainable}. Furthermore, innovative methods like delta-XAI, which leverages sensitivity analysis to rank features’ impact on outcomes, are being developed to enhance local explanations of predictions, thus increasing trust and satisfaction in AI-driven decisions \cite{de2024introducing}. Counterfactual explanations, which suggest minimal changes to achieve a different outcome, have also shown promise in improving user understanding and confidence, further bridging the gap between AI’s technical complexity and human interpretability \cite{warren2023categorical}.

In extending the principles of the D'Hondt method and proportional representation to AI model interpretability, it is conceivable that concepts like alliance systems and threshold applications—commonly used in real-world electoral processes—could inspire structured approaches to feature importance in complex machine learning models. For instance, just as electoral alliances allow smaller parties to gain representation by aligning with larger ones, AI models could adopt alliance-like groupings to evaluate how minor features contribute collectively alongside major features to influence predictions. This approach could help highlight nuanced interactions that might otherwise be overshadowed in traditional feature importance analyses, similar to how electoral alliances bring forward diverse voices that might not independently meet representation thresholds. Additionally, thresholds—like electoral cutoffs that parties must surpass to gain seats—could be applied in feature selection to filter out the least impactful variables, ensuring that only features with significant predictive influence are highlighted. This "thresholding" can enhance model clarity by focusing on the most informative aspects of data, improving interpretability without overwhelming users with less relevant details.

By applying these voting principles, such as the D'Hondt system's proportional allocation and alliance thresholds, to the interpretability of tree-based models, AI researchers can create a more nuanced ranking of feature importance that emphasizes both individual and collective feature contributions. This structured prioritization aligns with Explainable AI's goals, bridging model complexity with user understanding. Moreover, leveraging alliance systems and thresholds could offer valuable insights for managing feature importance in neural networks and ensemble models, where interpreting non-linear relationships is often challenging. Such methods can support greater transparency in high-stakes fields like healthcare and finance by ensuring that each feature's role in decision-making is proportionally represented and appropriately highlighted. Ultimately, incorporating real-world electoral principles into AI could pave the way for more interpretable, balanced, and democratic approaches to machine learning, enabling clearer, more accountable AI systems that resonate with human understanding and societal values. Here, a novel unified approach to interpreting model predictions is presented.
\footnote{\url{https://github.com/turkerbdonmez/dhondtxai}}

\section{Methodology}

The \texttt{DhondtXAI} library applies a structured methodology to evaluate feature importance in machine learning models, focusing on decision tree-based algorithms. It uses the D’Hondt proportional representation method to distribute importance fairly and transparently across features or feature groups. This method allows users to specify custom parameters, such as thresholds, exclusions, alliances, and the total numbers of votes and seats, enabling a flexible and insightful feature analysis.

\subsection{User-Defined Parameters}

The following parameters are defined by the user to establish the scope and constraints of the analysis:

\begin{itemize}
    \item \textbf{Total Number of Votes} (\( V \)): The user-specified total number of votes to be allocated proportionally among features based on their calculated importance values. This parameter enables direct control over the scale of the voting distribution.

    \item \textbf{Total Number of Seats} (\( M \)): The user-defined total number of seats (representative units) that will be allocated across features based on their final calculated vote shares, reflecting the overall distribution of influence among features.

    \item \textbf{Excluded Features} (\( F_{\text{exclude}} \)): A list of features the user wishes to remove from the analysis. These features will be excluded from all stages of the analysis, ensuring that they do not affect the importance-based voting or seating calculations.

    \item \textbf{Feature Alliances} (\textit{Alliances}): Specific groupings of features that the user defines to be evaluated together as a single unit. More than one alliance can be formed, and each alliance can include any number of features. For instance, if the user defines alliances such as \( \{ f_1, f_2 \} \), \( \{ f_3, f_4, f_5 \} \), and \( \{ f_6, f_9 \} \), each group is treated as a single alliance with a cumulative importance score. This score represents the combined importance of all features in the alliance, treated as a singular voting entity. The library is flexible in that alliances can contain overlapping or independent groups of features as specified by the user.

    \item \textbf{Threshold Value} (\textit{Threshold}): A minimum importance threshold defined as a percentage. Features or alliances whose relative importance is below this threshold are filtered out and do not receive any seats in the final seating allocation. This parameter allows the user to focus the analysis on only the most influential features or alliances.
\end{itemize}

These parameters create a customized analytical framework, ensuring that only the user-defined aspects of the feature set contribute to the voting and seating allocations.
\subsection{Calculation of Feature Importances in Tree-Based Models}

The foundation of the DhondtXAI process is the feature importance calculation, derived from tree-based models like Random Forests and Gradient Boosted Trees. In these models, feature importance is calculated by evaluating each feature’s contribution to reducing impurity at each decision node where the feature is used as a split criterion.

\paragraph{Reduction in Impurity at Node Level:} For any feature \( A \) used at a split node \( n \), the impurity before the split, \( I(n) \), and after the split, \( I'(n) \), are compared to assess the feature’s contribution to predictive improvement. The impurity after the split is a weighted sum of the impurities in the child nodes \( n_L \) and \( n_R \):

\begin{equation}
I'(n) = \frac{|n_L|}{|n|} I(n_L) + \frac{|n_R|}{|n|} I(n_R)
\end{equation}

where \( |n| \) is the number of samples in the parent node \( n \), and \( |n_L| \) and \( |n_R| \) represent the numbers of samples in the left and right child nodes, respectively.

\paragraph{Information Gain for Feature \( A \):} The reduction in impurity, or information gain \( \Delta I_{A,n} \), from splitting on feature \( A \) at node \( n \) is then calculated as:

\begin{equation}
\Delta I_{A,n} = I(n) - I'(n)
\end{equation}

This score reflects the extent to which feature \( A \) contributes to predicting the target variable at that specific split.

\paragraph{Aggregation Across Nodes and Trees:} Summing \( \Delta I_{A,n} \) over all nodes where feature \( A \) is used across all trees in the ensemble provides the cumulative importance score for that feature. For feature \( A \) in a single tree, the importance is:

\begin{equation}
\text{Importance}_{A, \text{tree}} = \sum_{n \in N_{A}} \Delta I_{A, n}
\end{equation}

where \( N_{A} \) is the set of nodes where \( A \) is used. Across all trees in an ensemble, the feature importance for \( A \) is averaged:

\begin{equation}
\text{Importance}_{A, \text{ensemble}} = \frac{1}{|T|} \sum_{t \in T} \text{Importance}_{A, t}
\end{equation}

Here, \( T \) represents the set of all trees, and \( \text{Importance}_{A, t} \) is feature \( A \)’s importance in tree \( t \).

These feature importance values form the basis for the proportional voting distribution used in the DhondtXAI methodology.

\subsection{Feature Exclusion and Alliance Formation}
\label{sec:step1}
The first step involves applying exclusions and forming alliances to establish the final feature set that will be analyzed.

\paragraph{Feature Exclusion:} Given the initial feature set \( F \), features specified by the user for exclusion are removed from the analysis. The resulting set of features used for voting and seat allocation, \( F' \), is defined as:

\begin{equation}
F' = F \setminus F_{\text{exclude}}
\end{equation}

where \( F_{\text{exclude}} \) denotes the set of features explicitly excluded by the user. Only features in \( F' \) are considered in subsequent steps, allowing for a refined analysis that adheres to user-defined constraints.

\paragraph{Alliance Formation:} Following feature exclusion, alliances are formed among features in \( F' \) based on user-defined groupings. Each alliance combines the importance values of its constituent features, enabling them to act as a unified feature. Multiple alliances are supported, each forming an independent entity that will participate in the voting and seat allocation process.

Let’s assume the user specifies several alliances: \( \{ f_1, f_2 \} \), \( \{ f_3, f_4, f_5 \} \), and \( \{ f_6, f_9 \} \). Each of these alliances is assigned a combined importance score based on the sum of the individual importance values of the features within the group.

The importance of each alliance \( \text{alliance}_j \) is calculated as follows:

\begin{equation}
\text{importance}_{\text{alliance}_j} = \sum_{i \in \text{alliance}_j} \text{importance}_{i}
\end{equation}

where \( \text{alliance}_j \) represents each unique alliance defined by the user, and \( i \) represents the individual features within that alliance.

Each alliance is treated as a single unit in the voting and seating calculations, meaning that it will receive votes and seats proportionate to its cumulative importance score rather than the individual scores of its constituent features. This cumulative score \( \text{importance}_{\text{alliance}_j} \) reflects the joint influence of all features within the alliance.

\paragraph{Finalized Feature and Alliance Sets:} After exclusions and alliances are processed, the final analysis set consists of:
\begin{itemize}
    \item Individual features not excluded or grouped into alliances.
    \item Each user-defined alliance, treated as a unified entity with a combined importance score.
\end{itemize}

These elements form the basis for the proportional vote and seat distribution, with each alliance’s total importance score influencing its share of votes relative to individual features.

\subsection{Initial Vote Distribution Using the D’Hondt Method}
\label{sec:step2}
With the final feature and alliance sets established, the initial distribution of votes is calculated based on the relative importance values of each feature or alliance.

\paragraph{Vote Distribution Across Features and Alliances:} Using the total votes \( V \) specified by the user, each feature or alliance \( i \) in the finalized set is allocated a proportional share of the total votes based on its importance score. The initial vote allocation for each feature or alliance \( i \) is calculated as follows:

\begin{equation}
\text{initial\_vote}_{i} = \frac{\text{importance}_{i}}{\sum_{j \in F'} \text{importance}_{j}} \times V
\end{equation}

where \( V \) represents the total votes, \( \text{importance}_{i} \) is the importance score for feature or alliance \( i \), and \( F' \) represents the final set of features and alliances after exclusions and groupings.

This formula ensures that features or alliances with higher importance scores receive a larger share of the votes, aligning the voting allocation directly with each feature’s or alliance’s calculated significance within the model.

\paragraph{Consideration for Multiple Alliances:} Since each alliance’s importance score is cumulative, the total importance score for alliances is the sum of the importance scores for all individual features within each alliance. This cumulative approach means that, despite multiple alliances or groupings, each alliance acts as a single voting entity.

For instance, if alliance \( \{ f_1, f_2 \} \) has a combined importance score of 0.30, and \( \{ f_3, f_4, f_5 \} \) has a combined score of 0.25, these alliances would receive initial votes proportionate to their total influence compared to other features or alliances.

\paragraph{Establishing the Vote Basis for Each Feature and Alliance:} The initial votes calculated for each feature or alliance \( \text{initial\_vote}_{i} \) serve as the foundation for further calculations. These initial votes directly influence both the threshold application (in section \ref{sec:step3}) and the final seat allocation process (in section \ref{sec:step4}), ensuring that every feature’s or alliance’s share of the total votes is transparently proportional to its relative importance within the model.

In this way, the initial vote distribution allows for a democratic and transparent reflection of each feature or alliance’s influence within the analysis framework, laying the groundwork for the remaining steps in the DhondtXAI process.

\subsection{Threshold Application and Redistribution of Votes}
\label{sec:step3}
The third step introduces the concept of a threshold—a user-defined minimum vote percentage that each feature or alliance must meet to be eligible for seat allocation. This threshold allows the user to filter out features or alliances with lower influence, ensuring that only those with significant importance values contribute to the final seat distribution.

\paragraph{Threshold Calculation:} The threshold value \( \text{threshold\_vote} \) is determined as a percentage of the total votes \( V \). This value acts as a minimum criterion that all features and alliances must meet to proceed to the seat allocation phase. Mathematically, the threshold vote amount is calculated as follows:

\begin{equation}
\text{threshold\_vote} = \frac{\text{threshold}}{100} \times V
\end{equation}

where \( \text{threshold} \) is the user-defined percentage (for example, 5\%).

If \( \text{threshold} = 0 \), this means there is no minimum vote requirement, allowing all features and alliances to move directly to the seat allocation phase. If \( \text{threshold} > 0 \), any feature or alliance whose initial vote count \( \text{initial\_vote}_{i} \) falls below \( \text{threshold\_vote} \) will be excluded from seat allocation, and their votes will be redistributed to the remaining features and alliances.

\paragraph{Identifying Below-Threshold Features and Alliances:} After calculating \( \text{threshold\_vote} \), each feature or alliance \( i \) is evaluated to see if it meets the threshold. If a feature’s or alliance’s initial vote count \( \text{initial\_vote}_{i} \) is less than \( \text{threshold\_vote} \), it is classified as a below-threshold entity and is excluded from receiving seats:

\begin{equation}
\text{below\_threshold} = \{ i \in F' : \text{initial\_vote}_{i} < \text{threshold\_vote} \}
\end{equation}

Features and alliances in this below-threshold group do not participate in the seat allocation phase, but their vote share is not disregarded; instead, it is redistributed to the remaining features and alliances.

\paragraph{Redistribution of Below-Threshold Votes:} The votes of all below-threshold features and alliances are collected and proportionally redistributed to the above-threshold group based on their relative importance. This redistribution step ensures that the votes of lower-importance features continue to contribute to the final allocation process, albeit indirectly, by strengthening the influence of more impactful features.

The total votes from below-threshold features, \( \sum_{l \in \text{below\_threshold}} \text{initial\_vote}_{l} \), is distributed across each feature \( j \) in the above-threshold group, proportional to each feature’s importance score:

\begin{equation}
\text{redistributed\_vote}_{j} = \text{initial\_vote}_{j} + \left( \frac{\text{importance}_{j}}{\sum_{k \in \text{above\_threshold}} \text{importance}_{k}} \times \sum_{l \in \text{below\_threshold}} \text{initial\_vote}_{l} \right)
\end{equation}

Here:
\begin{itemize}
    \item \( \text{redistributed\_vote}_{j} \) represents the final vote total for each feature or alliance in the above-threshold group after receiving redistributed votes.
    \item \( \text{importance}_{j} \) is the importance of the feature or alliance \( j \) within the above-threshold set.
    \item \( \sum_{k \in \text{above\_threshold}} \text{importance}_{k} \) normalizes the redistributed votes based on the importance scores of the remaining, eligible features.
\end{itemize}

This redistribution process provides a final vote count for each eligible feature or alliance, ensuring that even below-threshold features indirectly impact the results, thus preserving a fair representation of their cumulative influence.

\subsection{Seat Allocation Using the D’Hondt Method}
\label{sec:step4}
In the final step, seats are allocated to features and alliances based on their redistributed vote totals, using the D’Hondt method to proportionally distribute the specified total seats \( M \). This iterative process ensures that each feature or alliance receives seats in alignment with its importance and final vote count.

\paragraph{Initial Seat Ratios:} Each feature or alliance begins with an initial seat count of \( k = 0 \), meaning no seats have been allocated at the start. The D’Hondt method calculates a seat ratio \( S_i \) for each feature or alliance \( i \) based on its final vote count and current seat count. This initial ratio \( S_i \) is defined as:

\begin{equation}
S_i = \frac{\text{redistributed\_vote}_{i}}{k + 1}
\end{equation}

where:
\begin{itemize}
    \item \( \text{redistributed\_vote}_{i} \) represents the final vote count after redistribution for feature or alliance \( i \),
    \item \( k \) is the current number of seats allocated to \( i \) (initially \( k = 0 \)).
\end{itemize}

This initial seat ratio serves as the basis for the first iteration of seat allocation, with each feature or alliance’s eligibility for a seat determined by the highest current \( S_i \) value.

\paragraph{Iterative Seat Assignment:} The D’Hondt method proceeds iteratively, allocating one seat at a time based on the current seat ratios of all eligible features and alliances. In each iteration:
\begin{itemize}
    \item The feature or alliance with the highest \( S_i \) value receives one seat.
    \item Once a seat is allocated, the seat count \( k \) for the awarded feature or alliance is increased by 1, and its seat ratio \( S_i \) is recalculated to account for the updated seat count:
\end{itemize}

\begin{equation}
S_i = \frac{\text{redistributed\_vote}_{i}}{k + 1}
\end{equation}

This recalculated \( S_i \) adjusts the feature or alliance’s priority for receiving additional seats in subsequent iterations.

\paragraph{Completion of Seat Allocation:} The iterative seat assignment continues until all \( M \) seats have been allocated. Each time a seat is awarded, the \( S_i \) values are updated to reflect the new seat distribution, ensuring that seats are consistently distributed according to the highest remaining influence in each iteration.

\paragraph{Example of Seat Calculation in Practice:} Consider a feature with a final redistributed vote count of 1000 and an initial seat count \( k = 2 \). For this feature, the recalculated seat ratio \( S_i \) after receiving two seats would be:

\begin{equation}
S_i = \frac{1000}{2 + 1} = \frac{1000}{3} \approx 333.33
\end{equation}

This updated seat ratio is then compared to the \( S_i \) values of other features and alliances to determine the next recipient of a seat. By recalculating \( S_i \) at each step, the D’Hondt method ensures that seats are dynamically assigned based on the most current distribution of influence.

The seat allocation phase results in a final distribution of seats across features and alliances that reflects their relative importance, adhering to the principles of proportional representation. This approach allows users to clearly interpret the role of each feature or alliance within the model’s decision-making framework, providing a transparent and equitable view of feature influence.

\section{Results}
\subsection{Wisconsin Breast Cancer Dataset}
In this study, our focus is on comparing SHAP (SHapley Additive exPlanations) and DhondtXAI in assessing feature importance for a breast cancer classification model. Using the Wisconsin Breast Cancer Dataset \cite{breast_cancer_wisconsin_diagnostic_17} with only the mean feature values, we aim to demonstrate how each explainability technique identifies the contributions of individual features to the model's predictions. By contrasting SHAP and DhondtXAI, we can evaluate the interpretability and insights provided by each method, highlighting their effectiveness in explaining the model’s decision-making process in a medical context. This analysis will emphasize the practical value of explainable AI techniques in understanding model behavior and supporting diagnostic decision-making.

\begin{figure}[h]
  \centering
  \begin{subfigure}{0.45\textwidth}
    \includegraphics[width=\linewidth]{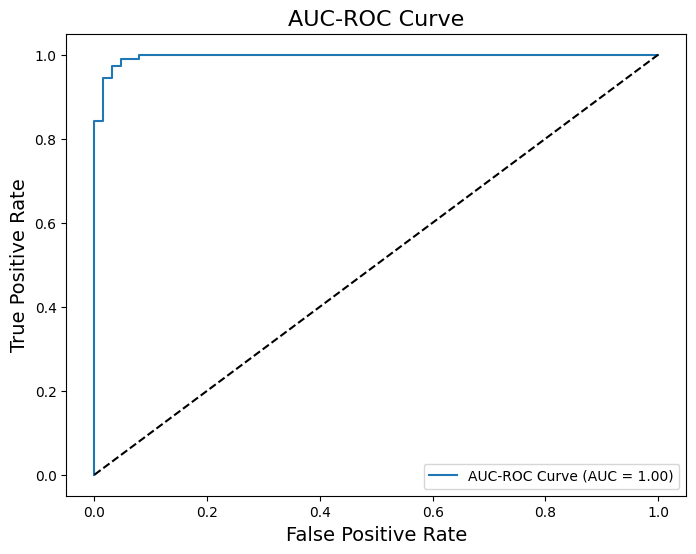}
    \caption{AUC-ROC Curve}
    \label{fig:aucroc}
  \end{subfigure}
  \hspace{0.05\textwidth}
  \begin{subfigure}{0.45\textwidth}
    \includegraphics[width=\linewidth]{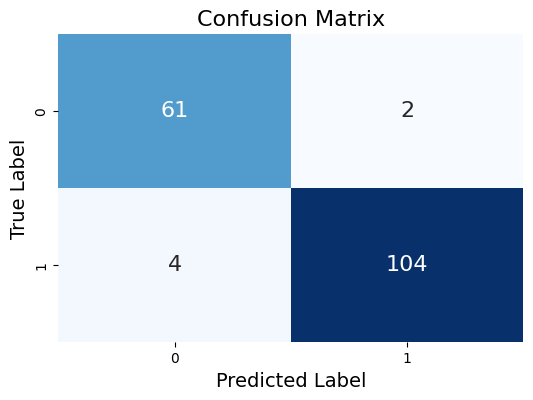}
    \caption{Confusion Matrix}
    \label{fig:confusionmatrix}
  \end{subfigure}
  \caption{Model Performance Metrics}
  \label{fig:performance}
\end{figure}

In this analysis, we utilized the Breast Cancer Wisconsin (Diagnostic) dataset to develop a machine learning model using only the mean features, reducing the dataset to 10 core attributes related to the mean measurements of cell nuclei. We applied the CatBoost classifier to this reduced feature set to build a model capable of distinguishing between benign and malignant tumors. The dataset, consisting of 569 observations, was split into training and testing sets, with 70\% allocated for training and 30\% for testing. The trained CatBoost model achieved an impressive accuracy of 96\%, with an F1 score of 0.97, recall of 0.96, and precision of 0.98. The AUC-ROC curve \ref{fig:aucroc}, with an AUC value of 1.00, demonstrated the model’s high effectiveness in class separation. Confusion matrix of the model also given in \ref{fig:confusionmatrix}

\subsubsection{Applying SHAP}
\begin{figure}[h]
  \centering
  \begin{subfigure}{0.45\textwidth}
    \includegraphics[width=\linewidth]{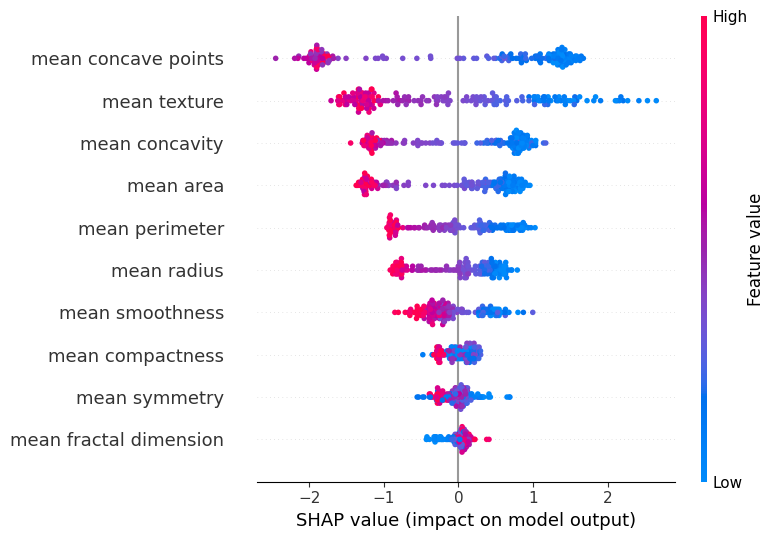}
    \caption{Detailed SHAP Summary Plot}
    \label{fig:summary}
  \end{subfigure}
  \hspace{0.05\textwidth}
  \begin{subfigure}{0.45\textwidth}
    \includegraphics[width=\linewidth]{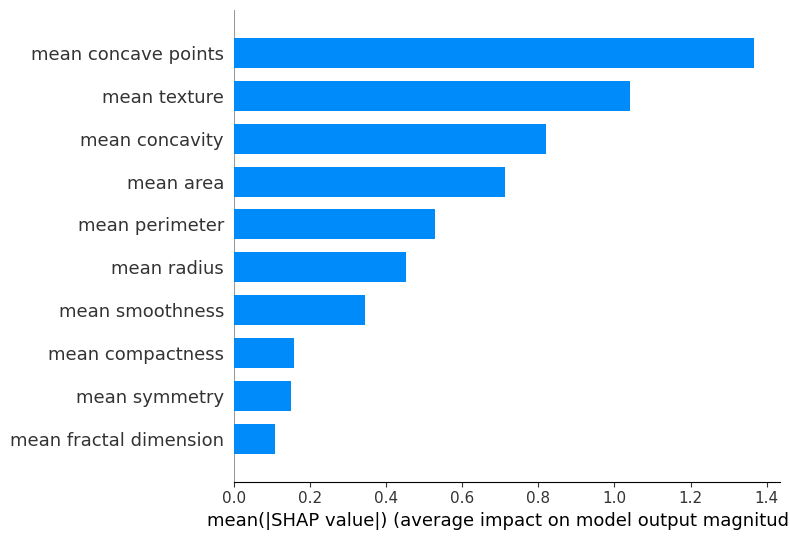}
    \caption{Global SHAP Summary Plot}
    \label{fig:globalshap}
  \end{subfigure}
  \caption{SHAP Analysis for Feature Importance}
  \label{fig:shap_analysis}
\end{figure}

SHAP (SHapley Additive exPlanations) was applied to the CatBoost model to gain insights into the impact of each feature on the model’s predictions. The first visualization, a SHAP summary plot(Figure \ref{fig:summary}, was generated to provide an overview of the distribution of SHAP values for each feature across all observations. Through this plot, features with the highest influence were identified, as well as how different feature values (high or low) affected the model output. "Mean concave points," "mean texture," and "mean concavity" were highlighted as highly impactful features, with their SHAP values showing the most significant effect on model predictions. In examining the direction of the SHAP values, it was observed that most features, including "mean concave points," "mean texture," and "mean concavity," were negatively correlated with the prediction of benign tumors, indicating that higher values for these features increase the likelihood of malignancy. In contrast, "mean fractal dimension" showed a positive correlation, where higher values were associated with a benign classification, suggesting that as fractal dimension increases, the likelihood of malignancy decreases.

Additionally, a global SHAP values bar plot (Figure \ref{fig:globalshap}) was created to represent the average magnitude of SHAP values for each feature, effectively summarizing each feature’s overall contribution to the model. In this plot, "mean concave points" was again shown to have the highest impact, followed by "mean texture" and "mean concavity," reinforcing their importance in predicting breast cancer diagnosis. Through these visualizations, a comprehensive understanding of feature importance was provided, allowing the model's decision-making process to be explained in a transparent and interpretable manner.

\subsubsection{Applying DhondtXAI}
In this analysis, the DhondtXAI method was utilized to allocate a 600-seat parliamentary representation based on the importance of each feature in the CatBoost model. This approach provided an intuitive and interpretable way of visualizing feature importance by representing each feature as a political party that receives "seats" according to its influence on the model’s predictions. Through this resource allocation, no variables were excluded, and no alliances were set, ensuring that each feature competed independently for representation. A total of 100,000,000 votes were distributed across the features, with each seat representing a proportionate share of influence as determined by the D’Hondt method.

\begin{figure}[h]
  \centering
  \includegraphics[width=\textwidth]{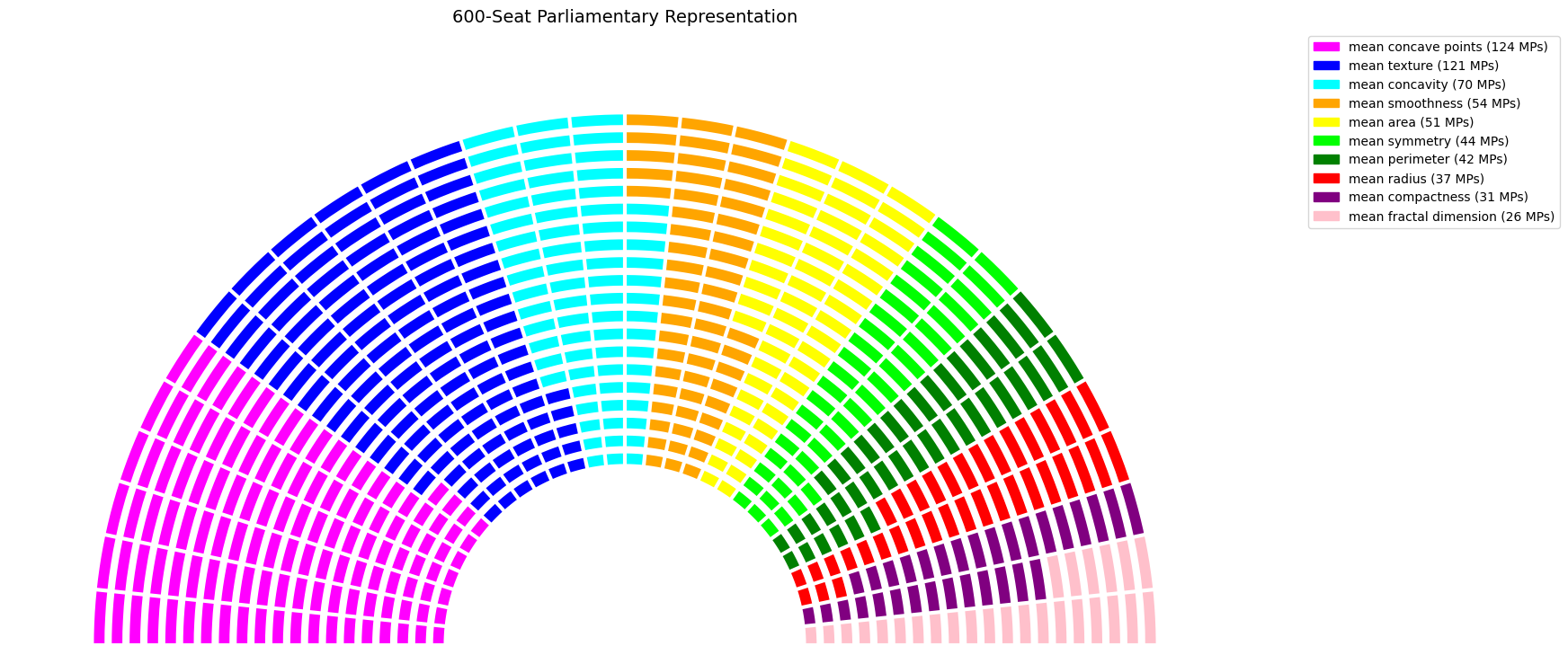}
  \caption{600-Seat Parliamentary Representation}
  \label{fig:parliament}
\end{figure}

The parliamentary representation view (Figure \ref{fig:parliament}) visualizes this allocation, where "mean concave points" stands out with the highest representation, occupying 124 seats. This feature’s strong influence suggests it is crucial for the model’s decision-making process in predicting cancer diagnosis. Following closely, "mean texture" received 121 seats, reinforcing its significance in classification. "Mean concavity," "mean smoothness," and "mean area" also held prominent positions with 70, 54, and 51 seats, respectively, indicating their substantial impact on model outcomes.

\begin{figure}[h]
  \centering
  \includegraphics[width=\textwidth]{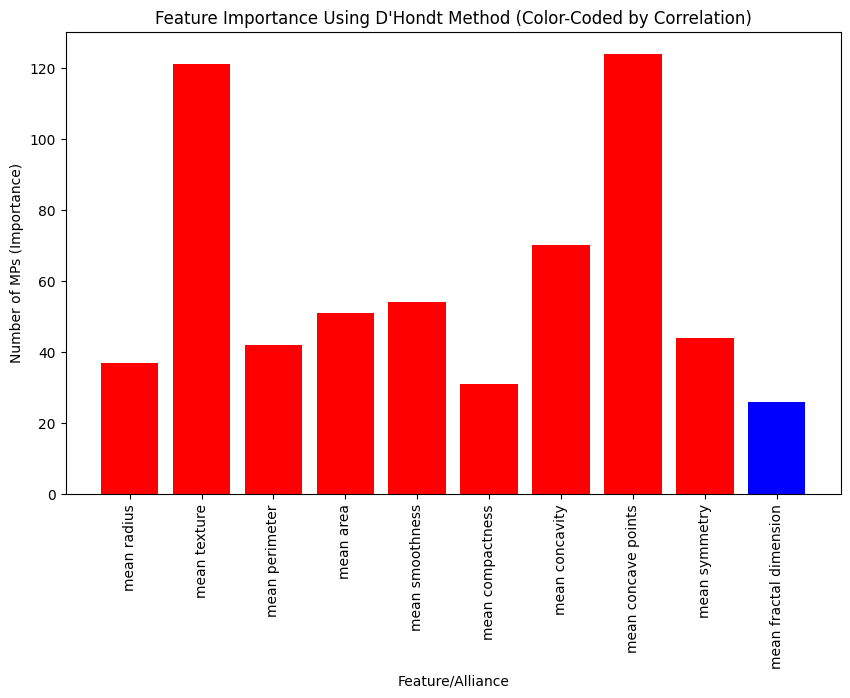}
  \caption{Bar-plot Represantation of MPs with correlation info}
  \label{fig:barplot}
\end{figure}

The color-coded seating arrangement showed in bar plot (Figure \ref{fig:barplot}) further highlights the correlation of each feature with the target outcome. Features marked in red, such as "mean texture," "mean concave points," and others, are negatively correlated, meaning higher values increase the likelihood of a malignant diagnosis. Conversely, "mean fractal dimension," represented in blue, is positively correlated, with higher values indicating a greater likelihood of benign classification. This contrast emphasizes the diverse roles features play within the model, showing not only their importance but also their directional impact.

The parliamentary view offers a unique and accessible visualization, enabling a comparative understanding of feature importance akin to political representation. Through this metaphorical representation, stakeholders can intuitively interpret the model’s internal mechanics, seeing how each feature contributes to the overall predictive performance in a visually structured and interpretable format. This analysis, supported by both the seat allocation and color-coded correlations, provides a comprehensive and transparent insight into the model’s feature dependencies, aiding in a deeper understanding of the predictive factors in breast cancer classification.

\subsubsection{Comparing SHAP and DhondtXAI}
In comparing SHAP and DhondtXAI, both methods provide insights into feature importance in the CatBoost model, yet from different interpretative angles. SHAP offers a direct calculation of each feature’s contribution to individual predictions, giving a global SHAP value that summarizes the average impact of each feature on model output. DhondtXAI, on the other hand, interprets feature importance through a resource allocation perspective, where each feature competes for “votes” that are converted into "MPs" in a parliamentary representation. This method not only ranks features by importance but also visualizes their influence as seats in parliament, enhancing interpretability.

Table \ref{tab:dhondt_xai} illustrates the outputs from both approaches, showing the votes, MP allocation, and global SHAP values for each feature. A clear alignment is observed between the two methods, with highly influential features, such as "mean concave points" and "mean texture," ranking at the top in both SHAP and DhondtXAI. Furthermore, the correlation direction between features and model output matches across the methods, supporting the consistency of the interpretations. For example, "mean concave points," "mean texture," and "mean concavity" display high importance in both SHAP values and DhondtXAI MP counts, while features with lower SHAP values, like "mean fractal dimension," also receive fewer MPs.
\begin{table}[h]
 \caption{D'Hondt XAI Vote Distribution, MPs in Parliament, and Global SHAP Values}
  \centering
  \begin{tabular}{lrrr}
    \toprule
    Feature & Votes & MPs in Parliament & Global SHAP Value \\
    \midrule
    mean radius & 6,271,130 & 37 & 0.451963 \\
    mean texture & 20,068,244 & 121 & 1.040490 \\
    mean perimeter & 7,038,052 & 42 & 0.529402 \\
    mean area & 8,600,885 & 51 & 0.713427 \\
    mean smoothness & 8,942,949 & 54 & 0.345782 \\
    mean compactness & 5,174,850 & 31 & 0.158138 \\
    mean concavity & 11,584,731 & 70 & 0.820642 \\
    mean concave points & 20,593,389 & 124 & 1.366414 \\
    mean symmetry & 7,344,121 & 44 & 0.151690 \\
    mean fractal dimension & 4,381,644 & 26 & 0.107822 \\
    \bottomrule
  \end{tabular}
  \label{tab:dhondt_xai}
\end{table}

A statistical test confirms the relationship between these two methods. A Spearman correlation between MP counts from DhondtXAI and global SHAP values yields a correlation coefficient of 0.83 with a p-value of 0.0029, indicating a statistically significant positive correlation (p < 0.05). This high correlation supports the agreement between SHAP and DhondtXAI in capturing feature importance, demonstrating that both methods provide reliable, complementary perspectives. Through this comparison, it becomes evident that DhondtXAI and SHAP together can offer robust insights, with DhondtXAI providing an intuitive, resource-based view that complements SHAP’s precise attribution of feature effects.

\subsection{Early Stage Diabetes Risk Prediction Dataset}
\begin{figure}[h]
  \centering
  \begin{subfigure}{0.45\textwidth}
    \includegraphics[width=\linewidth]{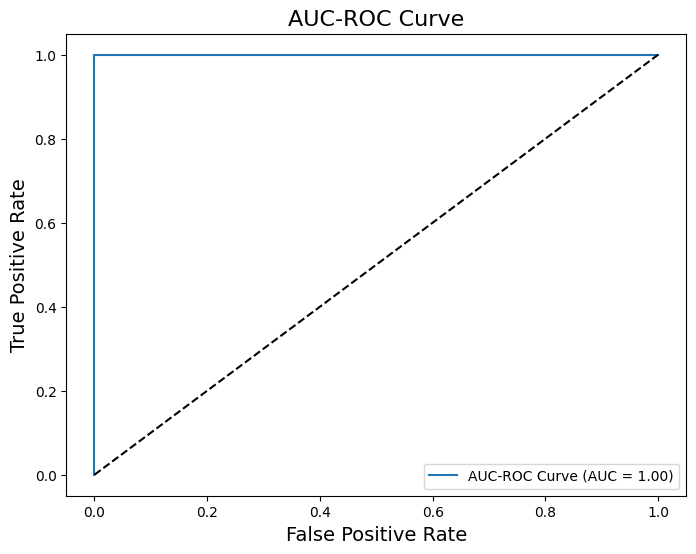}
    \caption{AUC-ROC Curve}
    \label{fig:aucroc2}
  \end{subfigure}
  \hspace{0.05\textwidth}
  \begin{subfigure}{0.45\textwidth}
    \includegraphics[width=\linewidth]{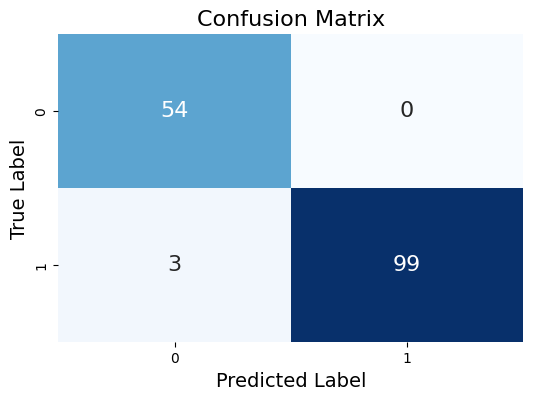}
    \caption{Confusion Matrix}
    \label{fig:confusionmatrix2}
  \end{subfigure}
  \caption{Model Performance Metrics}
  \label{fig:performance2}
\end{figure}

In this follow-up analysis, SHAP (SHapley Additive exPlanations) and DhondtXAI were applied to assess feature importance in an early-stage diabetes risk prediction model. The Early Stage Diabetes Risk Prediction dataset from the UCI Machine Learning Repository \cite{early_stage_diabetes_risk_prediction_529} was used, with binary and categorical variables converted into numerical formats to prepare the data for modeling. This dataset, which includes features such as age, gender, and symptoms like polyuria, polydipsia, and sudden weight loss, was utilized to predict diabetes risk. An XGBoost classifier was trained on this dataset, achieving a strong performance with an accuracy of 98\%, F1 score of 0.99, recall of 0.97, and precision of 1.00. The AUC-ROC curve, with an AUC value of 1.00, demonstrated excellent separation between classes, as reflected in the AUC-ROC plot (Figure \ref{fig:aucroc2}), while the confusion matrix illustrated the model’s ability to accurately classify diabetes risk (Figure \ref{fig:confusionmatrix2}).

As in the previous model, SHAP and DhondtXAI were employed to explore feature importance and interpretability. SHAP was used to reveal the direct impact of each feature on the model’s predictions, providing a global view of feature contributions. In contrast, DhondtXAI was applied to allocate feature importance using a resource-distribution approach, translating the influence of each feature into a parliamentary representation. This comparative analysis emphasizes the unique insights offered by both methods; SHAP highlights individual feature contributions, while DhondtXAI offers an intuitive, resource-based perspective that complements SHAP's detailed feature attribution. Through these explainability techniques, enhanced transparency is achieved, supporting interpretability and potentially aiding in clinical decision-making related to diabetes risk.

\subsubsection{Applying SHAP and Alliance-Based Feature Grouping}
\begin{figure}[h]
  \centering
  \begin{subfigure}{0.45\textwidth}
    \includegraphics[width=\linewidth]{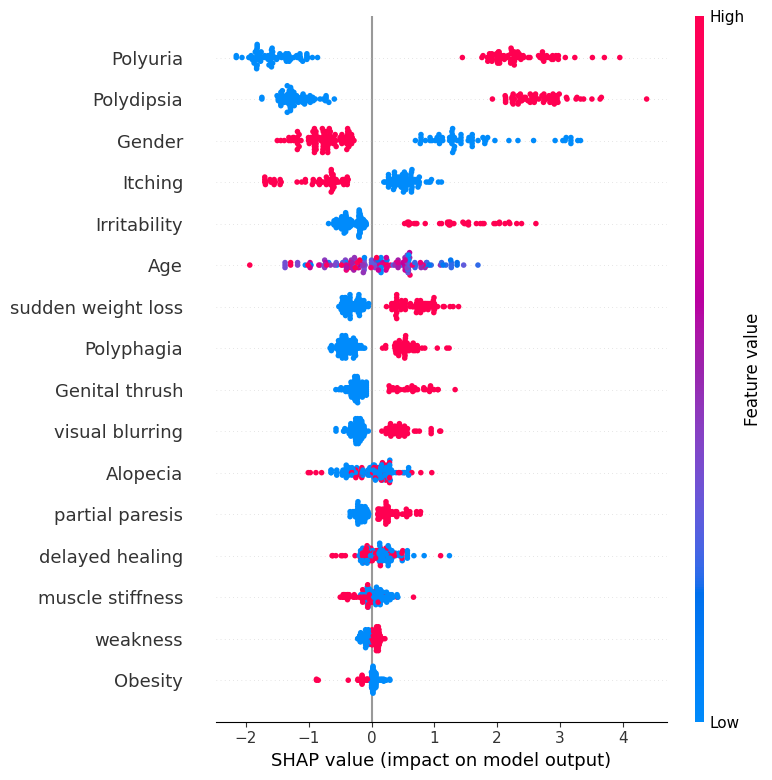}
    \caption{Detailed SHAP Summary Plot}
    \label{fig:summary2}
  \end{subfigure}
  \hspace{0.05\textwidth}
  \begin{subfigure}{0.45\textwidth}
    \includegraphics[width=\linewidth]{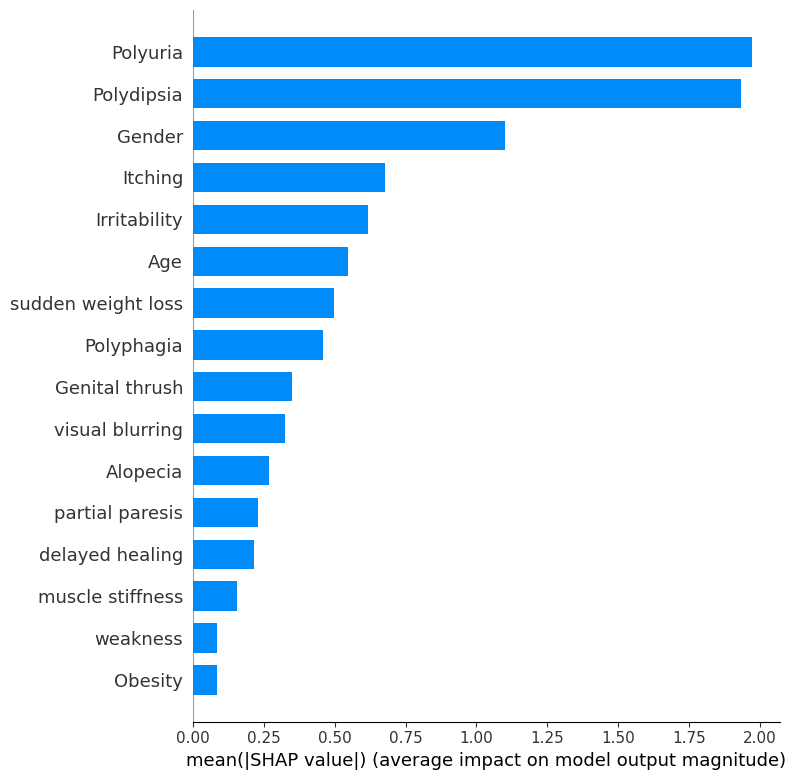}
    \caption{Global SHAP Summary Plot}
    \label{fig:globalshap2}
  \end{subfigure}
  \caption{SHAP Analysis for Feature Importance}
  \label{fig:shap_analysis2}
\end{figure}

SHAP (SHapley Additive exPlanations) was applied to the XGBoost model to gain insights into the impact of each feature on the model’s predictions. A SHAP summary plot (Figure \ref{fig:summary2}) was generated to provide an overview of the distribution of SHAP values for each feature across all observations. Through this plot, the most influential features were identified, as well as how different feature values (high or low) affected the model output.

In this analysis, "Polyuria", "Polydipsia", and "Gender" emerged as the features with the most significant impact on the predictions, as indicated by their SHAP values. It was observed that higher values of "Polyuria" and "Polydipsia" positively influenced the likelihood of a diabetes prediction, meaning that increased levels of these features are associated with a higher probability of diabetes. For "Gender" (coded as 1 for male and 0 for female), being male showed a positive association with the likelihood of a diabetes prediction, indicating that males have a higher probability of a positive diagnosis according to the model. Conversely, features such as "Obesity" and "weakness" exhibited a relatively lower impact and did not strongly affect the model's prediction of diabetes.

Examining the direction of the SHAP values revealed further insights. "Polyuria", "Polydipsia", and "Gender" (where being male is associated with a higher probability of diabetes) showed a positive impact, indicating that higher values of these features increase the likelihood of a positive diabetes prediction. On the other hand, certain features such as "visual blurring" and "Alopecia" exhibited a negative correlation with diabetes predictions, suggesting that higher values for these features decrease the likelihood of a positive diagnosis.

Additionally, a global SHAP values bar plot (Figure \ref{fig:globalshap2}) was produced to represent the average magnitude of SHAP values for each feature, effectively summarizing the overall contribution of each feature to the model. This plot reinforced the importance of "Polyuria", "Polydipsia", and "Gender" in predicting diabetes, as these features were shown to have the highest impact. Through these visualizations, a comprehensive understanding of feature importance was provided, facilitating an interpretable and transparent explanation of the model’s decision-making process."

In order to enhance interpretability of the model’s predictions, features were organized into alliances, which are groups representing related aspects of diabetes risk factors and symptoms. Each alliance was constructed based on the logical association of features, allowing us to view the model's predictions through broader thematic categories rather than individual variables. SHAP (SHapley Additive exPlanations) values were then computed for each alliance by calculating the mean of the absolute SHAP values for the features within each group. This approach offers a summarized view of feature importance at the alliance level, making the model's decision-making process more interpretable and cohesive.

The following alliances were defined for this analysis:

\begin{itemize}
    \item \textbf{Metabolic\_Body}: This alliance includes \texttt{Age}, \texttt{Gender}, and \texttt{Obesity}, representing demographic and physiological factors that are known to influence general health and body composition. Age and gender are fundamental demographic factors, while obesity is a significant risk factor for diabetes, often associated with metabolic health. Together, these features provide insights into the patient’s baseline health status.

    \item \textbf{Diabetes\_Symptoms}: Composed of \texttt{Polyuria}, \texttt{Polydipsia}, \texttt{sudden weight loss}, and \texttt{Polyphagia}, this alliance represents the classic symptoms of diabetes. These symptoms are commonly observed in diabetic patients due to the body’s inability to manage blood glucose levels effectively. By grouping these features, we capture a holistic view of symptomatic indicators of diabetes, making this alliance critical for prediction.

    \item \textbf{Skin\_Infection}: This alliance consists of \texttt{Itching}, \texttt{Alopecia}, \texttt{Genital thrush}, and \texttt{delayed healing}, reflecting common skin and infection-related issues seen in individuals with diabetes. High blood glucose levels can impair immune function and skin health, leading to slower healing and susceptibility to infections. Grouping these features highlights the indirect yet relevant effects of diabetes on skin and immune health.

    \item \textbf{Vision\_Nervous}: This alliance includes \texttt{visual blurring}, \texttt{partial paresis}, \texttt{muscle stiffness}, and \texttt{weakness}, representing neurological and vision-related issues that may result from prolonged diabetes. Diabetic neuropathy and other nervous system issues can arise from poorly managed blood glucose, leading to symptoms affecting muscle and nerve function as well as vision. This group encapsulates the broader neurological impacts of diabetes.

    \item \textbf{Psychological}: This alliance, represented by \texttt{Irritability}, reflects psychological and behavioral symptoms associated with diabetes. Changes in mood and behavior are often reported among diabetic patients, and irritability can stem from blood glucose fluctuations, affecting mental well-being.
\end{itemize}

\begin{figure}[h]
  \centering
  \includegraphics[width=\textwidth]{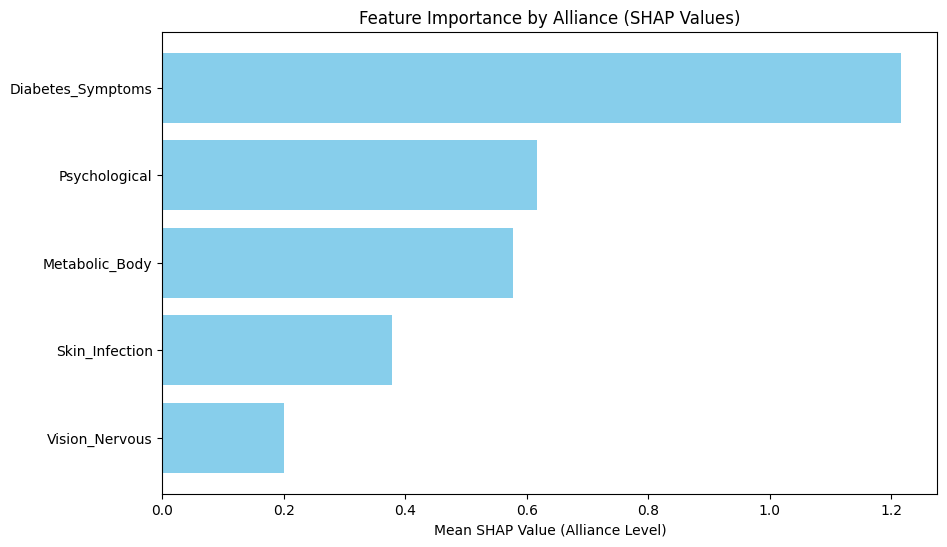}
  \caption{Alliance-based feature importance in diabetes prediction, showing mean SHAP values for grouped features, with the highest impact from classic diabetes symptoms.}
  \label{fig:feature_alliance}
\end{figure}

After grouping these features into alliances, the SHAP values were calculated at the alliance level, as displayed in \textbf{Figure \ref{fig:feature_alliance}}. In Figure \ref{fig:feature_alliance}, the bar plot demonstrates the relative importance of each alliance in predicting diabetes, based on the mean SHAP values. \textbf{Diabetes\_Symptoms} emerged as the most influential alliance, with a mean SHAP value of (1.2156). This high value suggests that the classic symptoms of diabetes have the strongest impact on the model’s predictions, aligning with the medical understanding of these symptoms as primary indicators of diabetes. The \textbf{Psychological} alliance, with a SHAP value of (0.6160), also contributed significantly, indicating that psychological factors, while secondary, have a notable association with diabetes predictions.

The \textbf{Metabolic\_Body} alliance, with a SHAP value of (0.5780), was also impactful, underscoring the importance of demographic and physiological factors such as age, gender, and obesity. In comparison, the \textbf{Skin\_Infection} alliance (0.3778) and \textbf{Vision\_Nervous} alliance (0.1995) showed lower SHAP values, suggesting that while these features are relevant, they have a relatively minor impact on the model’s predictions of diabetes.

By structuring the SHAP analysis around alliances, we obtain a clearer, more interpretable view of how groups of related features influence the model's decision-making process. This approach not only highlights individual feature importance but also emphasizes the combined effect of feature groups, providing a broader understanding of diabetes prediction factors as they relate to the XGBoost model.

\subsubsection{DhondtXAI and Alliance Feature}
In this analysis, the DhondtXAI method was applied to assess feature importance within a diabetes prediction model, focusing on the collective influence of logically grouped features, known as alliances. Initially, the process involved prompting the user to input any variables to exclude from the evaluation and to define alliances among the features. This setup enabled a tailored approach to feature grouping, where related features were combined into alliances to assess their aggregate impact on the model's predictions.

Once the alliances were defined, a total of 100 million votes and 600 parliamentary seats were allocated to simulate the D'Hondt method’s distribution mechanism. This method is particularly suited for this type of analysis, as it efficiently distributes influence across feature groups according to their relative importance, rather than focusing on individual features in isolation. The DhondtXAI method enhances interpretability by providing insights into how sets of related features collectively impact the model, aligning with a more holistic understanding of feature importance.

\begin{figure}[h]
  \centering
  \includegraphics[width=\textwidth]{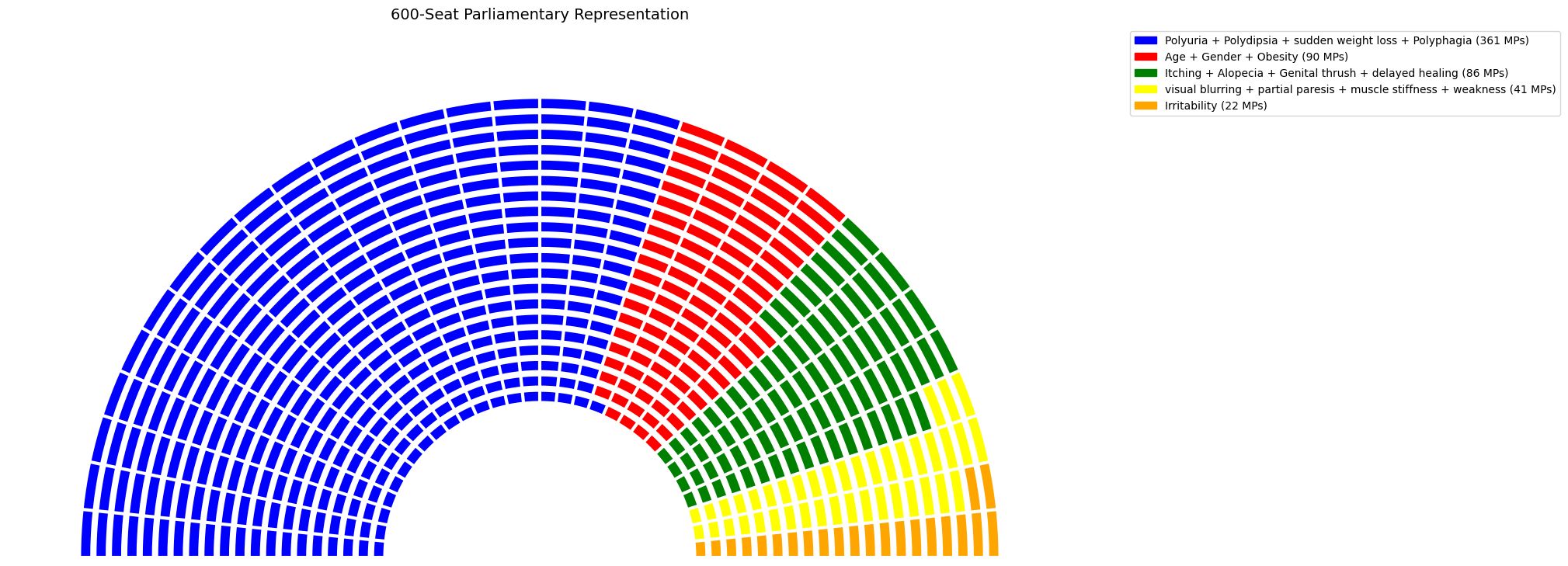}
  \caption{Bar-plot Represantation of MPs with Alliances with correlation info}
  \label{fig:parliament2}
\end{figure}

\begin{figure}[h]
  \centering
  \includegraphics[width=0.9\textwidth]{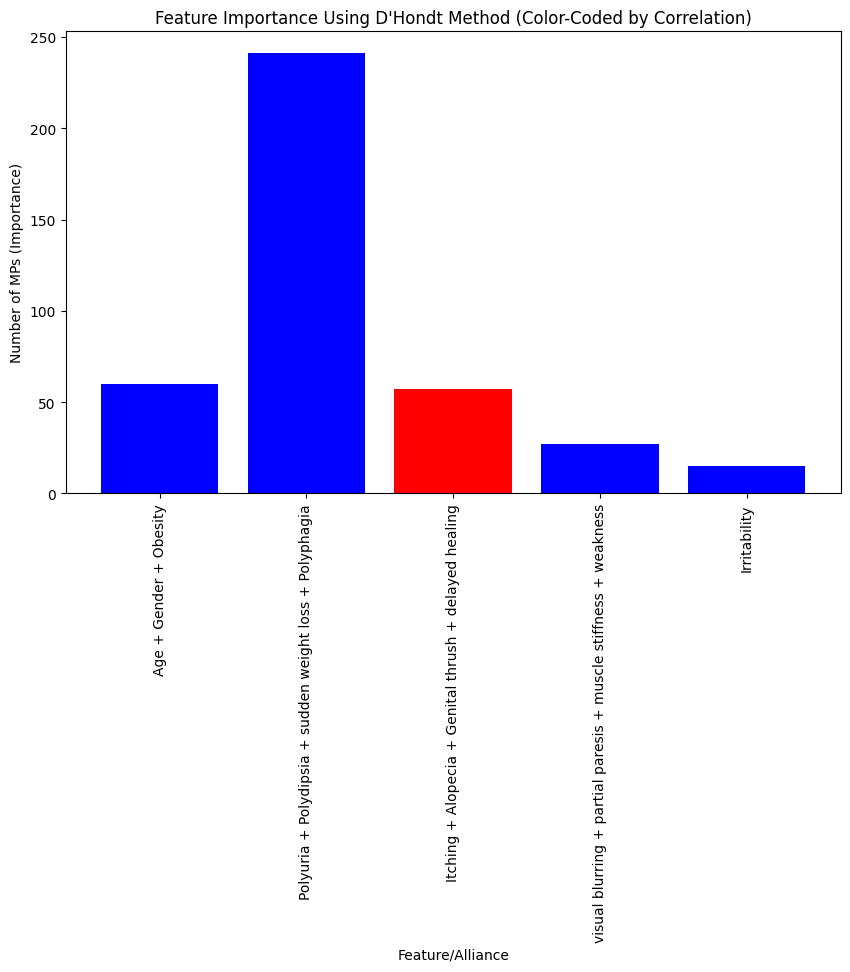}
  \caption{Bar-plot Represantation of MPs with Alliances with correlation info}
  \label{fig:barplot2}
\end{figure}

The results of this analysis are visualized in \textbf{Figure \ref{fig:parliament2}} and \textbf{Figure \ref{fig:barplot2}}. \textbf{Figure \ref{fig:parliament2}} presents a 600-seat parliamentary representation of the model's feature importance. In this circular plot, each alliance occupies a portion proportional to its relative importance in the model’s decision-making process. The \textbf{Diabetes\_Symptoms} alliance, encompassing classic diabetes indicators such as \texttt{Polyuria} and \texttt{Polydipsia}, occupies the largest segment with 361 seats, underscoring its dominant influence on the model’s predictions. Following this, the \textbf{Metabolic\_Body} alliance, which groups demographic and physiological factors like \texttt{Age}, \texttt{Gender}, and \texttt{Obesity}, was allocated 90 seats, reflecting its substantial, though secondary, role. Other alliances, including \textbf{Skin\_Infection} with 86 seats and \textbf{Vision\_Nervous} with 41 seats, represent relevant but smaller contributions. Lastly, the \textbf{Psychological} alliance, represented solely by \texttt{Irritability}, occupies 22 seats, highlighting its smaller but still notable impact.

\textbf{Figure \ref{fig:barplot2}} provides a bar plot illustrating the importance of each alliance, with color coding to indicate their correlation with the prediction outcome. In this plot, blue bars represent alliances with a positive correlation to the likelihood of a positive diabetes diagnosis, indicating that higher values in these features increase the likelihood of a diabetes prediction. Conversely, red bars indicate a negative correlation, where higher feature values decrease the likelihood of a positive prediction. This color coding allows us to quickly discern not only the importance of each alliance but also the direction of their influence on the model’s predictions.

Together, these visualizations in Figures \ref{fig:parliament2} and \ref{fig:barplot2} provide a transparent and interpretable view of how alliances of related features influence the model’s decisions. By aggregating votes and distributing seats based on the importance of each alliance, the DhondtXAI method translates feature relevance into a format that is easily interpretable. This approach not only highlights which alliances are most influential but also conveys the relative weight each alliance contributes to the model’s predictions, offering a nuanced and comprehensive understanding of the model’s behavior.
\subsubsection{Comparing SHAP and DhondtXAI}
In comparing SHAP and DhondtXAI, both methods provide insights into feature importance in the CatBoost model, yet from different interpretative angles. SHAP offers a direct calculation of each feature’s contribution to individual predictions, giving a global SHAP value that summarizes the average impact of each feature on model output. DhondtXAI, on the other hand, interprets feature importance through a resource allocation perspective, where each feature or alliance competes for “votes” that are converted into "MPs" in a parliamentary representation. This method not only ranks features or alliances by importance but also visualizes their influence as seats in parliament, enhancing interpretability.

Table \ref{tab:dhondt_xai_alliances} illustrates the outputs from both approaches, showing the votes, MP allocation, and global SHAP values for each alliance. A general alignment is observed between the two methods, with highly influential alliances, such as \textbf{Diabetes\_Symptoms} and \textbf{Metabolic\_Body}, ranking at the top in both SHAP and DhondtXAI. The \textbf{Diabetes\_Symptoms} alliance, for example, received the highest allocation with 361 MPs and had the highest SHAP value of 1.2156. Similarly, \textbf{Metabolic\_Body} received 90 MPs and a corresponding SHAP value of 0.5780, underscoring its importance in the model’s predictions.

\begin{table}[h]
 \caption{D'Hondt XAI Vote Distribution, MPs in Parliament, and Global SHAP Values}
  \centering
  \begin{tabular}{lrrr}
    \toprule
    Alliances & Votes & MPs in Parliament & Global SHAP Value \\
    \midrule
    Metabolic\_Body & 14,950,619 & 90 & 0.5780 \\
    Diabetes\_Symptoms & 60,007,287 & 361 & 1.2156 \\
    Skin\_Infection & 14,392,427 & 86 & 0.3778 \\
    Vision\_Nervous & 6,851,943 & 41 & 0.1995 \\
    Psychological & 3,797,721 & 22 & 0.6160 \\
    \bottomrule
  \end{tabular}
  \label{tab:dhondt_xai_alliances}
\end{table}

The results of a Spearman correlation analysis between the MP counts from DhondtXAI and global SHAP values yielded a correlation coefficient of 0.40 and a p-value of 0.505, suggesting a weak positive relationship that is not statistically significant (p > 0.05). While the correlation did not reach statistical significance, DhondtXAI offers a different perspective on feature importance, providing a more intuitive, alliance-based approach. In particular, the parliamentary visualization created by DhondtXAI presents a straightforward and interpretable view of each alliance’s relative importance, making it potentially more accessible and effective for end users who may benefit from a tangible representation of feature influence.

Through this comparison, it becomes evident that DhondtXAI and SHAP together can offer complementary insights. DhondtXAI provides an intuitive, resource-based perspective that complements SHAP’s precise attribution of feature effects. Despite the slight lack of significance in correlation, DhondtXAI’s parliamentary representation and alliance-based view of feature importance contribute meaningfully to understanding model behavior, especially in cases where grouped features or alliances are of interest.

\subsubsection{Threshold application over DhondtXAI}
The threshold application in DhondtXAI introduces a practical mechanism to filter out variables or alliances that have a minimal impact on the model, enabling a focused and refined interpretation of feature importance. In this analysis, a 10\% threshold was applied, meaning only alliances receiving at least 10\% of the total votes would be eligible for seat allocation. By implementing this threshold, DhondtXAI ensures that only the most influential feature groups contribute to the final interpretation, effectively reducing noise and highlighting the primary variables that affect the model's predictions.

The results are summarized in \textbf{Table \ref{tab:threshold_mp_distribution}}, which illustrates the change in the number of MPs for each alliance after applying the threshold. This adjustment led to an increase in MPs for alliances like \texttt{Metabolic\_Body}, \texttt{Diabetes\_Symptoms}, and \texttt{Skin\_Infection}, as their vote shares exceeded the threshold, securing 100, 404, and 96 MPs, respectively. However, alliances such as \texttt{Vision\_Nervous} and \texttt{Psychological} failed to meet the threshold requirement, resulting in a reduction to zero MPs. This outcome highlights the relative insignificance of these alliances under the current model setup, suggesting that their influence on the predictive output is limited compared to the more impactful groups.

\begin{table}[ht]
 \caption{Change in MP Allocation After Applying a 10\% Threshold}
  \centering
  \begin{tabular}{lrrr}
    \toprule
    Alliances & Distributed MPs & Distributed MPs with 10\% Threshold & Change in MPs Count \\
    \midrule
    Metabolic\_Body & 90 & 100 & +10 \\
    Diabetes\_Symptoms & 361 & 404 & +43 \\
    Skin\_Infection & 86 & 96 & +10 \\
    Vision\_Nervous & 41 & 0 & -41 \\
    Psychological & 22 & 0 & -22 \\
    \bottomrule
  \end{tabular}
  \label{tab:threshold_mp_distribution}
\end{table}

\begin{figure}[h]
  \centering
  \includegraphics[width=0.9\textwidth]{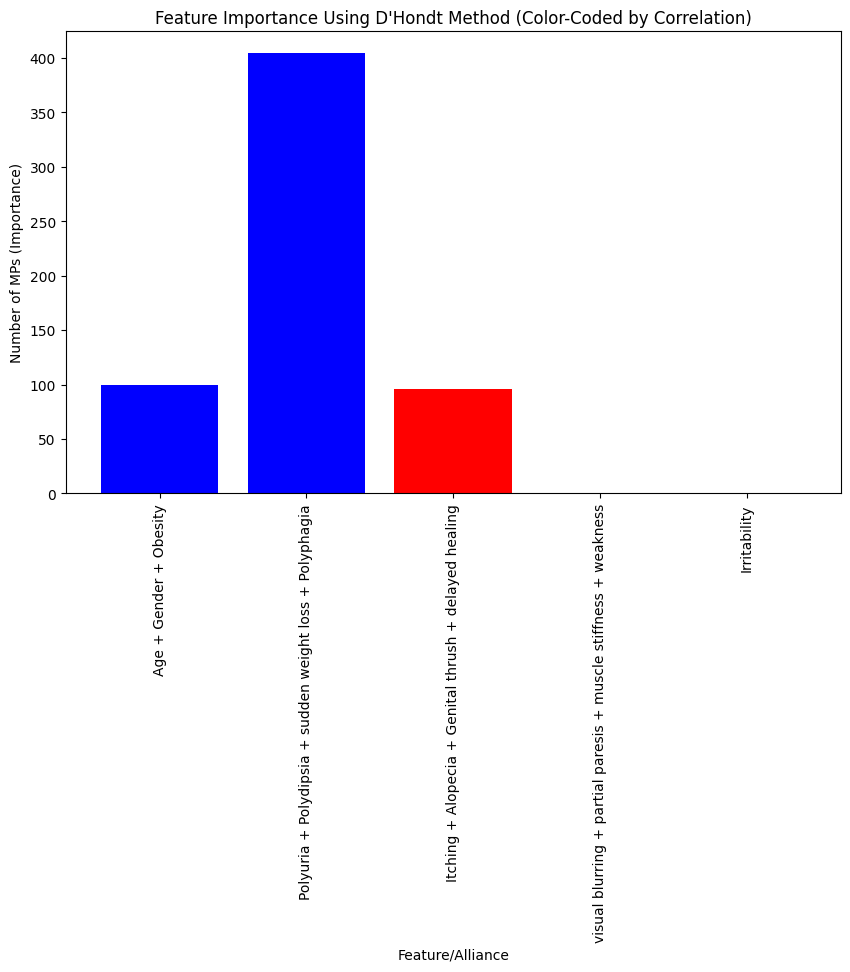}
  \caption{MP Distribution with 10\% Threshold and Correlation Indicators (Red = Negative Correlation, Blue = Positive Correlation)}
  \label{fig:mp_distribution_correlation}
  \FloatBarrier
\end{figure}

\textbf{Figure \ref{fig:mp_distribution_correlation}} provides a visual representation of the final allocation of MPs after threshold application, with color-coding to indicate positive and negative correlations with the target outcome. Alliances marked in red indicate a negative correlation, meaning higher values increase the likelihood of a particular model outcome (e.g., disease risk), while those in blue represent a positive correlation, suggesting higher values decrease the likelihood of this outcome. This distinction further enhances the interpretative clarity provided by DhondtXAI, allowing stakeholders to understand both the influence and directional impact of each alliance on the model's predictions.

By setting a threshold, DhondtXAI enables users to concentrate on the core variables that drive the model’s decision-making process, streamlining the interpretation and allowing for a clearer understanding of the main factors at play. In this context, the threshold not only reduces complexity but also reinforces the interpretative clarity of the analysis, focusing on key alliances that have a substantial role in the model's outcomes. Such a threshold mechanism is particularly valuable in applications where interpretability is essential, as it ensures that only the most relevant features are highlighted in the decision-making framework, thereby facilitating a more effective communication of the model’s insights to stakeholders.

Through this threshold application, DhondtXAI aligns with Explainable AI's goal of enhancing transparency and relevance in feature importance analysis. The allocation of MPs based on the threshold-adjusted vote distribution provides an accessible and intuitive way to interpret feature significance, allowing users to focus on the main predictive drivers without being overwhelmed by less impactful variables. This approach ultimately strengthens the utility of DhondtXAI as a tool for interpretable machine learning, particularly in high-stakes fields where clear and reliable insights are essential.

\section{Discussion}
\par In analyzing feature importance using both SHAP and DhondtXAI, this study provides a unique perspective on the interpretability of machine learning models by adapting a democratic principle to the evaluation of feature contributions. The integration of a democratic method, like the D'Hondt system, within the realm of Explainable AI (XAI) offers a structured approach that not only highlights individual feature importance but also accounts for alliances and thresholds, which can represent feature groups or combined influences. This form of structured prioritization mirrors real-world electoral systems where representation is allocated proportionally, ensuring a balance of influence that could otherwise be dominated by a few features. This democratic allocation may help address potential biases in feature importance by emphasizing collective influence, a valuable addition to the realm of interpretable machine learning models.

One of the strengths of using DhondtXAI in combination with traditional methods like SHAP lies in its ability to enhance interpretability in an intuitive, visually accessible way. DhondtXAI, by representing feature influence as parliamentary seats, allows users to perceive model behavior in a format that resonates with societal principles of fair representation. This not only aids experts but also provides a framework accessible to non-technical stakeholders, enabling them to grasp the relative contributions of feature groups in a machine learning model. The application of the D'Hondt method, therefore, democratizes model interpretation by giving each feature—or alliance of features—a voice in the decision-making process, much like electoral systems that seek to represent diverse societal groups.

The addition of alliances and thresholds in DhondtXAI also brings a layer of flexibility that traditional feature importance techniques may lack. Alliances allow users to explore the collective impact of feature groups, which can be particularly useful in domains where interconnected variables operate together, such as healthcare or social sciences. In this context, alliances make it possible to assess how minor features, which may have limited standalone influence, can contribute meaningfully when considered as part of a group. This is akin to smaller political parties forming alliances to gain parliamentary representation, thus making it a relevant analogy for interpreting complex machine learning models. Moreover, thresholds enable the exclusion of less influential features, focusing attention on those with the most significant impact and improving clarity for the end-user.

A comparative analysis of SHAP and DhondtXAI results in this study further demonstrates the reliability and consistency of both methods in identifying influential features. Although the two approaches differ in their methodologies—SHAP providing a more granular, mathematical attribution of feature contributions, and DhondtXAI focusing on a proportional distribution model—the outcomes are largely consistent, especially for features with high importance. This consistency is statistically supported by a strong positive Spearman correlation, underscoring the complementary nature of the two methods. While SHAP excels in detailed, precise explanations of individual feature impact, DhondtXAI offers a more intuitive, aggregate view that aligns well with democratic principles of proportional representation.

In considering the democratic nature of DhondtXAI, it also opens the door to ethical and socially responsible AI practices. The proportional representation framework encourages fairness and transparency, which are essential qualities for AI models used in sensitive fields like healthcare, finance, and criminal justice. By ensuring that each feature or group of features is adequately represented according to its influence, DhondtXAI mitigates the risk of feature bias, a common challenge in AI. This approach aligns with the broader goals of Ethical AI, fostering a system that values interpretability and inclusivity. In a time when AI systems are increasingly scrutinized for their transparency and accountability, incorporating democratic methods like DhondtXAI could enhance user trust and confidence in AI-driven decisions.

However, while DhondtXAI offers notable advantages, it is essential to recognize that this method may not be universally applicable across all types of models and datasets. The effectiveness of DhondtXAI depends on the structure of the data and the relationships between features, as well as the context in which interpretability is sought. For instance, alliance-based groupings may be less relevant in domains where features are largely independent or lack clear interaction effects. Additionally, for models that require highly granular insights into individual predictions, traditional XAI methods such as SHAP may remain more appropriate due to their precise attributions. Nevertheless, in domains where group interactions are crucial, DhondtXAI provides an innovative, context-sensitive tool that complements traditional XAI techniques.

In conclusion, the use of DhondtXAI to assess feature importance through a democratic lens exemplifies an innovative intersection of political theory and machine learning interpretability. By leveraging the D'Hondt method to proportionally allocate influence among features or alliances, DhondtXAI introduces a unique, accessible approach to understanding complex models. The democratic principles embedded in this approach provide a balanced view of feature influence, fostering transparency, fairness, and accountability in AI applications. This method not only complements traditional XAI techniques like SHAP but also opens new pathways for ethical AI practices, supporting a future where AI interpretability aligns closely with societal values. The DhondtXAI method could, therefore, serve as a valuable addition to the toolkit of Explainable AI, particularly in high-stakes environments where democratic representation of feature importance is essential for building trust and understanding.
\section{Conclusion}
In this study, we explored the application of the D’Hondt method, a well-established proportional representation technique, to enhance interpretability in machine learning models through the DhondtXAI library. By adapting electoral principles such as seat allocation and alliance formation to feature importance in AI models, DhondtXAI provides a novel and intuitive framework for explaining complex machine learning outputs. This method allows features or groups of features, treated as "parties" or "alliances," to "compete" for "seats" or representational units based on their calculated importance, thereby offering a structured, transparent way of distributing interpretative emphasis across model features.

DhondtXAI’s framework aligns with recent advancements in Explainable AI (XAI) by prioritizing features proportionally according to their contributions, which enhances user understanding and trust in the model's decisions. The parliamentary view, with its seat allocation visualization, makes the often complex distribution of feature importance accessible, intuitive, and engaging for non-technical stakeholders. This alignment with democratic principles in AI design has significant potential to bridge the interpretability gap, making complex models not only more understandable but also more accountable.

Furthermore, the threshold application feature in DhondtXAI allows for the exclusion of less impactful features, focusing interpretive resources on the most influential variables. This mirrors the electoral process of excluding minor parties that do not meet a specified threshold, thus streamlining decision-making by concentrating on the dominant players. This feature can be particularly valuable in domains such as healthcare, finance, and legal systems, where only the most impactful variables should influence critical decisions.

The concept of democratizing AI through DhondtXAI has broader implications for ethical AI development. By incorporating democratic principles like proportional representation and alliances, DhondtXAI lays a foundation for models that represent a fair distribution of feature importance, much like a democratic parliament reflects a fair distribution of societal preferences. This can lead to AI systems that are more aligned with human values, ensuring that each feature’s role in a model is represented in a balanced, interpretable, and fair manner.

In practice, DhondtXAI can empower stakeholders to better understand and engage with AI-driven decisions, fostering transparency and accountability. The method’s potential for application across various sectors highlights its flexibility as a tool for democratizing AI interpretability. As AI continues to play a growing role in society, approaches like DhondtXAI represent a significant step toward creating AI systems that are not only technically robust but also socially responsible and transparent.

\bibliographystyle{unsrt}  

\end{document}